\documentclass[pdflatex,sn-mathphys]{sn-jnl}%
\usepackage{ amssymb }
\usepackage{indentfirst}
\usepackage{tabularx}
\jyear{2022}%
\title{ModMag: A Modular Magnetic Micro-robotic Manipulation Device}
\usepackage{float}
\begin{document}
\author[1]{\fnm{Max} \sur{Sokolich}}\email{sokolich@udel.edu}
\author[1]{\fnm{David} \sur{Rivas}}\email{drivas@udel.edu}
\author[1]{\fnm{Markos} \sur{Duey}}\email{markosd@udel.edu}
\author[1]{\fnm{Daniel} \sur{Borsykowsky}}\email{dborsy@udel.edu}
\author[1]{\fnm{Sambeeta} \sur{Das*}}\email{samdas@udel.edu}
\affil[1]{\orgdiv{Department of Mechanical Engineering}, \orgname{University of Delaware}, \orgaddress{\street{130 Academy St}, \city{Newark}, \postcode{19717}, \state{DE}, \country{USA}}}

\normalsize

\abstract{Electromagnetic systems have been used extensively for the control magnetically actuated objects, such as in microrheology and microrobotics research. Therefore, optimizing the design of such systems is highly desired. Some of the features that are lacking in most current designs are compactness, portability, and versatility. Portability is especially relevant for biomedical applications in which in vivo or in vitro testing may be conducted in locations away from the laboratory microscope. This document describes the design, fabrication and implementation of a compact, low cost, versatile, and user friendly device (the ModMag) capable of controlling multiple electromagnetic setups, including a two-dimensional 4-coil traditional configuration, a 3-dimensional Helmholtz configuration, and a 3-dimensional magnetic tweezer configuration. All electronics and circuitry for powering the systems is contained in a compact 10"x6"x3" system which includes a 10" touchscreen. A graphical user interface provides additional ease of use. The system can also be controlled remotely, allowing for more flexibility and the ability to interface with other software running on the remote computer such as propriety camera software. Aside from the software and circuitry, we also describe the design of the electromagnetic coil setups and provide examples of the use of the ModMag in experiments.
}

\maketitle

\begin{table}[h]
\begin{center}
\begin{minipage}{\textwidth}
\caption{Specifications table}\label{tab1}%
\begin{tabular}{@{}ll@{}}

\toprule
Subject Area & Mechanical Engineering: Micro-Robotics \\
More Specific Subject Area    & Magnetic Manipulation Systems   \\
Name of Your Method         &  Modular Magnetic Manipulation System   \\
Name and Reference of Original Method    & Not Applicable   \\
Resource Availability              & CAD Files and Software Available Upon Request. \\

\botrule
\end{tabular}
\end{minipage}
\end{center}
\end{table}

\section{Introduction}
\subsection{Background}
Micro-robots have drawn a lot of attention in the past few years due to their potential to radically improve the efficacy of numerous tasks, such as cell manipulation, targeted drug delivery, microsurgery, and mixing of particles \cite{peyer2013bio}. Despite this, conventional methods for powering macro-scale robotic actuation methods due not apply at the micro-scale,since they cannot be scaled down to size. Hence, autonomous and reliable control of these micro-robots is an on going challenge. Different methodologies have been proposed in recent years in a bid to efficiently control micro robots. Researchers have validated the actuation of the micro machines employing bubble creation using chemical reactions \cite{solovev2010catalytic} or by attaching the synthetic micro-robots to swimming bacteria \cite{sakar2011modeling}. External magnetic, acoustic, or electric fields have also been employed to control and steer the micro-robots \cite{sakar2011modeling,aghakhani2020acoustically, das2018experiments}. 

One of the most common methods of controlling micro-robots is magnetic actuation. Magnetic actuation has multiple advantages over other actuation methods; for example, the micro-robots can be controlled independent of line of sight making them capable of deployment in variety of environments. In addition, magnetic fields are regarded as a safe choice to use at the cellular and tissue level for a large number of biomedical applications \cite{fischer2011magnetically}. 


The examples listed above all use expensive equipment such as data acquisitions devices and large bench top power supplies which limit their ability to be used on other microscopes or transported to other locations. Furthermore, the software is generally integrated onto a lab computer which also limits portability.

In this paper, we demonstrate the design and testing of a handheld, portable, low cost, modular electromagnetic actuation system, ModMag, for control of micro-robots. We also show experimental examples of micro-robot control using different types of magnetically driven micro-robots for applications in extracellular research.

\subsection{Electromagnetism}
When an electric current passes through a wire, the movement of charge generates a magnetic field.  A solenoid, or electromagnet, uses many of these wires wrapped concentrically around a cylinder to create a magnetic field in a direction along its central axis. The strength of the magnetic field is proportional to the current and increases with the number of times the wire is wrapped around the cylinder. It is also proportional to the magnetic permeability which can be increased by introducing certain materials into the core of the solenoid. Generally, ferromagnetic materials with low coercivity, such as soft iron, are the material of choice for their ability to enhance the magnetic field generated by the solenoid while retaining low magnetization when the current is turned off.

Ferromagnetic materials contain microscopic magnetic domains where groups of magnetic moments naturally align in the same direction. When the domains are all aligned randomly the material is demagnetized. However, the presence of an external magnetic field can align the domains in the direction of the applied magnetic field, causing the material to be magnetized. It may be helpful to think of each of the magnetic domains as tiny magnets; when current passes through the surrounding coil of wire, it forces each of these tiny magnets in the same direction, thereby amplifying the field. 

In a similar manner, when a ferromagnetic object, such as a magnetic micro-robot, is placed in a magnetic field, its magnetic moment aligns with the external field. This is the fundamental way in which magnetic microrobots are controlled. Consider a self propelled, spherical, iron coated micro-sphere suspended in a liquid solution. Typically, the micro-sphere moves in an arbitrary direction and is subject to Brownian rotational fluctuations. An external torque can be applied in order to steer the microsphere in a particular direction. The torque, $\boldsymbol{\Gamma}$, is given by

\begin{equation}
    \boldsymbol{\Gamma} = \textbf{m}\times\textbf{B}
\end{equation}

\noindent where $\textbf{m}$ is the magnetic moment of the microrobot and $\textbf{B}$ is the magnetic field. This torque therefore allows one to magnetically orient the microrobots. Magnetic forces can also be applied to the microrobots by applying magnetic gradients. The magnetic force, $\textbf{F}$, is given by

\begin{equation}
    \textbf{F} = \left(\textbf{m}\cdot\nabla\right)\textbf{B}
\end{equation}

\begin{figure}[H]
    \centering
    \includegraphics[width=12cm]{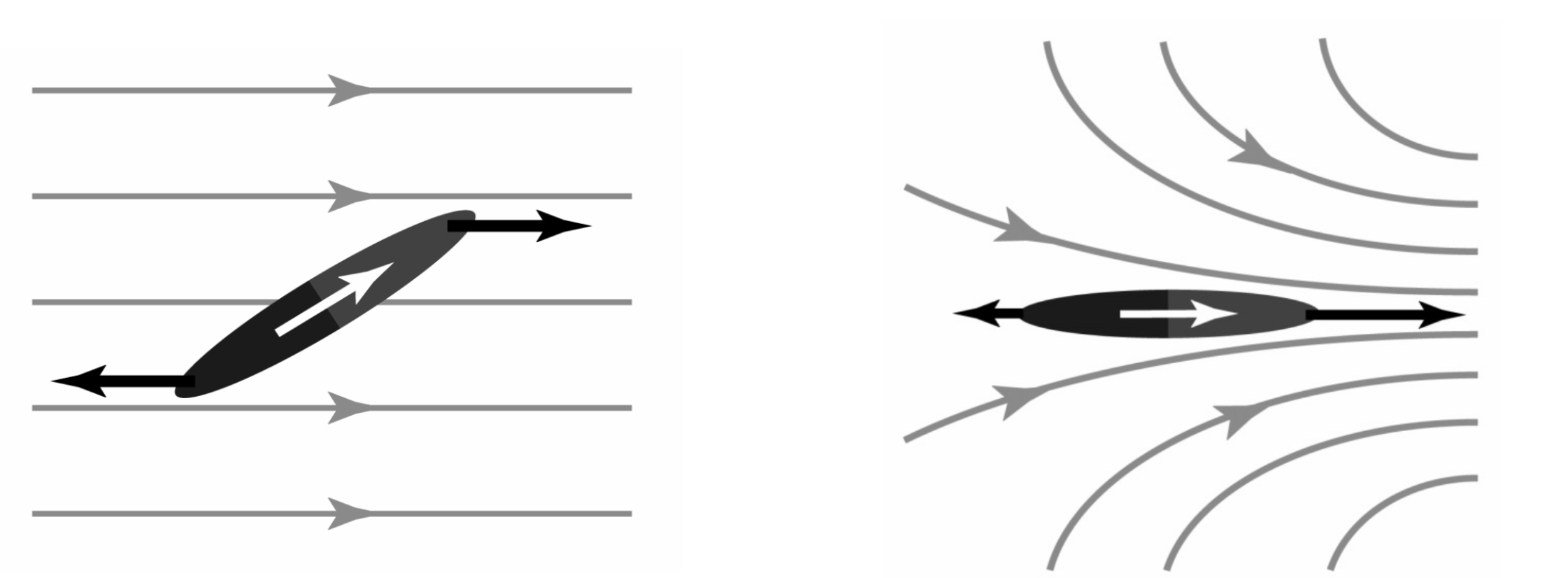}
    \caption{The left figure shows an object rotating due to a magnetic torque. The white arrow indicates the object's magnetic moment and the gray lines represent magnetic field lines. The figure on the right shows an object moving due to a magnetic force created by a magnetic field gradient \cite{vries_high_2004}. The region to the right of the object has a greater density of field lines which corresponds to a higher magnetic field strength.}
    \label{fig:Field Illustration}
\end{figure}

Applying sufficiently strong magnetic gradients generally requires the use of high permeability poles which are used to concentrate the magnetic flux generated from an electromagnetic. This allows for strong enough magnetic gradient forces to overcome the highly viscous drag forces that a microrobot experiences. Magnetic devices that are used to magnetically manipulate objects are known as magnetic tweezers. Figure \ref{fig:Field Illustration} illustrates the process by which an object with a magnetic moment can be oriented or translated in a uniform or spatially varying field, respectively.

\section{Method Details}
\subsection{Mechanical}
In determining the design of the portable electromagnetic system we focused on three aspects. The first was that the software and circuitry should be versatile in its ability to be used with multiple electromagnetic setups and, therefore, microrobots with different actuation mechanisms. That is, it is desirable for a system to be versatile and not require a complete system redesign when using different microrobots or during different applications or experiments. The electromagnetic setups we focused on were a 2D solenoid array, a 3D Helmholtz-based, and a 3D magnetic tweezer system. Each of these systems allows for immense versatility when it comes to micro-robotic experiments. We also sought to create a portable and inexpensive system that could also be used easily. Making the system portable and using a standalone design allows for the system to be transferable to different  microscopes and imaging systems and also to settings outside the lab which could be useful in medical applications. 

As described above, three designs were used: a traditional 2D coil system, a 3D Helmholtz-based coil system, and a 3D magnetic tweezer system. The designs of each are discussed in the following sections.

\subsubsection{2D Traditional Coils}
\begin{figure}[H]
    \centering
    \includegraphics[width=12cm]{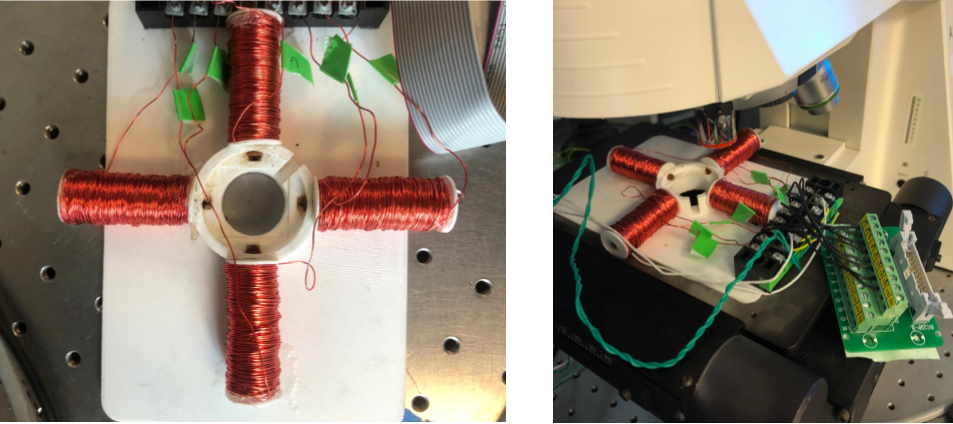}
    \caption{2D electromagnetic stand}
    \label{fig:2D}
\end{figure}

The traditional 2D electromagnetic setup consists of a 3D printed rectangular exchangeable stand that fits inside the microscope slide holder, and 4 perpendicular electromagnets (see Fig.~\ref{fig:2D}). The stand was designed such that it fit onto the mechanical stage of a Zeiss Axioplan 2 upright microscope, however it can be placed on any microscope stage surface. The Axioplan 2’s limited stage space of approximately 150 x 100 x 25 cm in volume made it difficult to design a stand with sufficient magnetic strength and negligible disruption of the microscope's functionality.  As a result both the stand and the electromagnetic coils were designed to be as compact as possible.

The 2D electromagnetic stand was designed in Solidworks and consists of only one part (see Figure \ref{fig:2D}). The height of the stand measures roughly 23 mm to accommodate for the rotation of all microscope objectives in the objective turret. The region of interest was designed such that a standard 22 x 22 mm cover-slip or 25 x 25 mm square glass slide could be loaded with the desired samples roughly half way up the height of the stand. In addition, a 30 mm petri dish fits perfect in the region of interest. This allows for the electromagnets to point directly towards the samples. In addition, a custom square-shaped holder avoids the need to use a microscope slide which would require the electromagnets to rest at an angle. A slit in the stand was introduced to allow for easy insertion and removal of the cover slip using laboratory tweezers. The electromagnetic stand was printed on a Comgrow Creality Ender 3 Pro 3D printer using 1.75 mm, white PLA filament. An STL file of the stand was uploaded to Creality’s Cura software with the following settings: printing temperature: 200 C, bed temperature 60 C, speed: 50 mm/s, layer height: 0.12 mm, and infill: 20. The rest of the settings were left as default. These settings provide a strong and durable stand for the electromagnets to rest upon. 4 equally spaced holes 5 mm in diameter were created along the same plane as the cover slip holder allowing for the insertion of each electromagnet core. The holes allow for the electromagnetic core to be as close to the cover slip as possible.

The electromagnetic core material is AISI 1008 carbon steel made of 99.31-99.7$\%$ iron, cold drawn and annealed at 925°C. The cores had raw dimensions of 5 mm in diameter by 200 mm long. The relative permeability of the core was listed as (86 $\pm$ 3.61) H/m. The cores were then cut to 50 mm long. 24 AWG magnet wire was used for the electromagnetic coils. In order to prevent the coil from unraveling, plastic end caps of 30 mm in diameter were cut from 0.7 mm in thickness plastic. 5 mm holes were then drilled into the center of these circles to allow for the electromagnetic core to be inserted. To create an electromagnet, one end cap is positioned at the end of the core and glued in place using Gorilla 2 part epoxy. The second end cap is then glued 50 mm from the first end cap, or 5 mm from the opposite side. As a result, the electromagnetic coil will be 50 mm long, leaving 5 mm at one side to be inserted into the stand. Coils were then wound carefully using an electric drill to exactly 980 turns of the 24 AWG magnet wire. Using a Tesla meter, the electromagnetic strength at the face of the coil was measured to be (201 $\pm$ 3) mT at a current of 2 Amps, and around 15 mT at the center of the work space. The six electromagnet's were then inserted and glued into the six holes in the 3D printed stand.

\subsubsection{3D Helmholtz-Based System}

\begin{figure}[H]
    \centering
    \includegraphics[width=12cm]{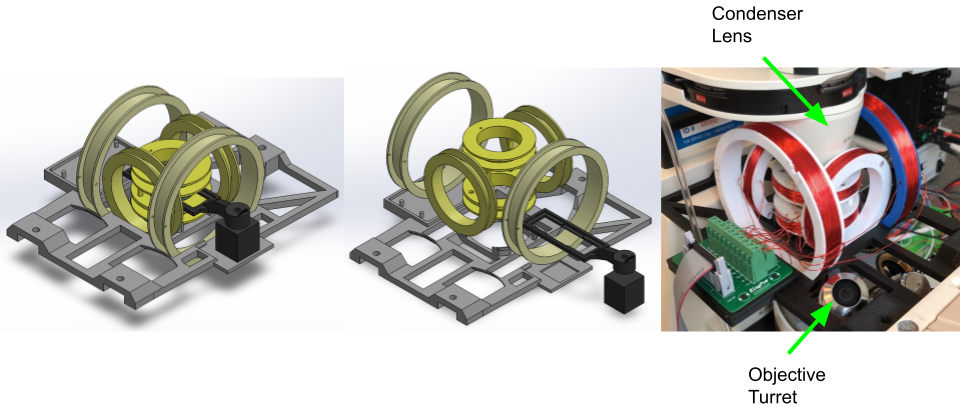}
    \caption{Design schematics and image of the 3D Helmholtz-based System}
    \label{fig:Helmholtz}
\end{figure}

A 3D electromagnetic manipulation system was designed for micro-robotic applications involving rotating homogeneous magnetic fields, which is well suited for 3D magnetic steering or magnetic rolling. The custom coil system is inspired by a  Helmholtz design and is designed for mounting on the Zeiss Axiovert 200 M inverted microscope (see Fig.~\ref{fig:Helmholtz}). Measurements of the microscope bed dimensions and distances between lenses were recorded. These size constraints then defined the maximum radii for the coil rings. Calipers were used to determine the microscope measurements. The Helmholtz coil system consists of three pairs of rings each sharing their respective common axis. The rings are wrapped with copper wire to form coils. As mentioned above, to attain an optimally uniform field the distance between pairs of rings should be equal to their radius. However, this is difficult to achieve in practice and is also generally unnecessary for microrobotic applications, therefore we loosened this constraint in order to allow the system to better fit the workspace. As we will discuss later, we find that this system performs very well for magnetic steering and rolling applications for which it is designed. We do not notice any effects of non-uniformity in any of our experiments.   

To optimally fit in the workspace, the large and medium sized rings were mounted on the stage vertically while the small rings were mounted horizontally, as shown in Fig.~\ref{fig:Helmholtz}. Microscope dimensions allow for a maximum diameter of 106 mm for the large rings without interfering with the objective turret, and a minimum diameter of 36 mm for the small rings without interfering with the condenser lens.  The smallest pair of rings are spaced such that their distance, $H_s$, is equal to their radius, $H_s = R =36 mm$. The medium sized pair is spaced slightly farther apart ($H_m = 66 mm$). For the largest pair of rings, $H_l = 83 mm$. As a result, the ratios of distance over radius of the small, medium, and large pairs of rings are 1.3, 1.8, and 1.5, respectively. The platform unit is roughly 230 mm long and 140 mm wide with integrated slots for mounting on the microscope. On either side of the platform is an area for mounting a motherboard with electrical connections, and a square divot for mounting a 2-axis micrometer for slide arm micro-translation. The slide arm fits both a conventional glass microscope slide as well as square cover-slips. Four slots are outfitted within the platform to hold the large and medium coils with a transition fit. The small coil pair sits on four stilts, while four small pegs separate the coil pair. 

The platform, slide arm, and rings were designed within Solidworks CAD software and printed using 3D Printing methods. All components are printed with PLA using an Ender 3 Pro. Structural properties of PLA allow for adequate rigidity within the frame to hold the weight of the primary components. The coils are wrapped with 24 AWG copper wire using a standard hand drill to achieve more uniform wire layers than what is achieved by hand. In order to connect the ring to the drill, a drill adapter unit consisting of three extending limbs was fabricated. Small pegs on each limb allowed for a transition fit into receiving holes on the rings. The resulting number of turns is approximately 368 for the small and medium coils and 260 for the large coils. Applying 1 A to each of the coils yields a uniform magnetic field over the sample area of 4 mT, 2 mT and 2 mT for the small, medium and large rings respectively. This is more than adequate for actuating non-gradient based magnetic micro-robots as shown in the experimental validation section.

\subsubsection{3D Magnetic Tweezers}
The magnetic tweezer system is inspired by the design presented in \cite{zhang20193d}. It consists of 6 magnetic poles made out of high permeability MuMetal. A high magnetic permeability material is necessary for the construction of the poles in order to attain efficient transport of the magnetic flux to the tips of the poles. The system has an easy to use substrate holder in which a microscope slide can be inserted into the workspace.

\begin{figure}[H]
    \centering
    \includegraphics[width=12cm]{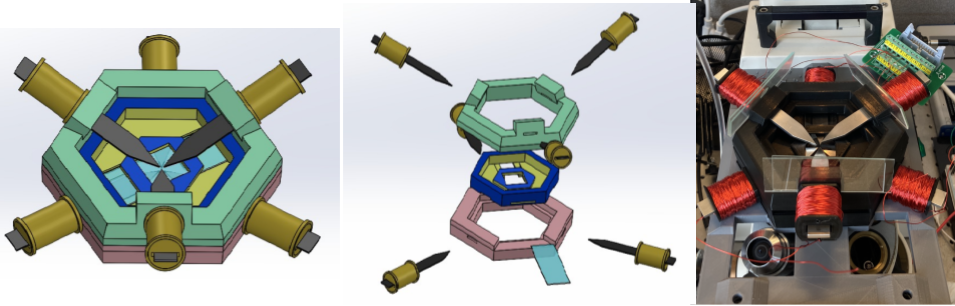}
    \caption{3D magnetic tweezer system based on \cite{zhang20193d}}
    \label{fig:Tweezer}
\end{figure}

This 3D design of magnetic tweezers allows for a concentrated and uniform magnetic field gradient surrounding the microrobots. The 3 poles coming from the top and bottom allow for magnetic symmetry as well as 3D maneuverability of magnetic particles, something not achievable with in-plane tweezers. With 6 sources of magnetic flux, actuation can be made more precise, and movement more robust.

The design aims to maximize field strength with large pole to pole tweezer distances as well as minimal air gaps for better conduction. The bottom hexagon orients the 3 bottom tweezers such that they point toward the bottom center of the slide (1.5mm separated), and also supports the middle hexagon, the slide holder in a transition fit. The top hexagon orients the top 3 tweezers in the same manner and fits around the slide holder. The simplicity of the case design allows for convenient assembly/disassembly, as well as no usage of hardware or adhesive. The case, including the coil/tweezers holders was 3D printed entirely out of PLA and the tweezers were cut from 1.5mm Mumetal, a nickel-iron soft ferromagnetic alloy with very high permeability. 

The limiting design constraint of these systems was space availability due to the microscope itself, and therefore the design may be altered or enhanced based on one's specific space constraints.

\subsection{Electronics}
\subsubsection{Components}

\begin{figure}[H]
    \centering
    \includegraphics[width=10cm]{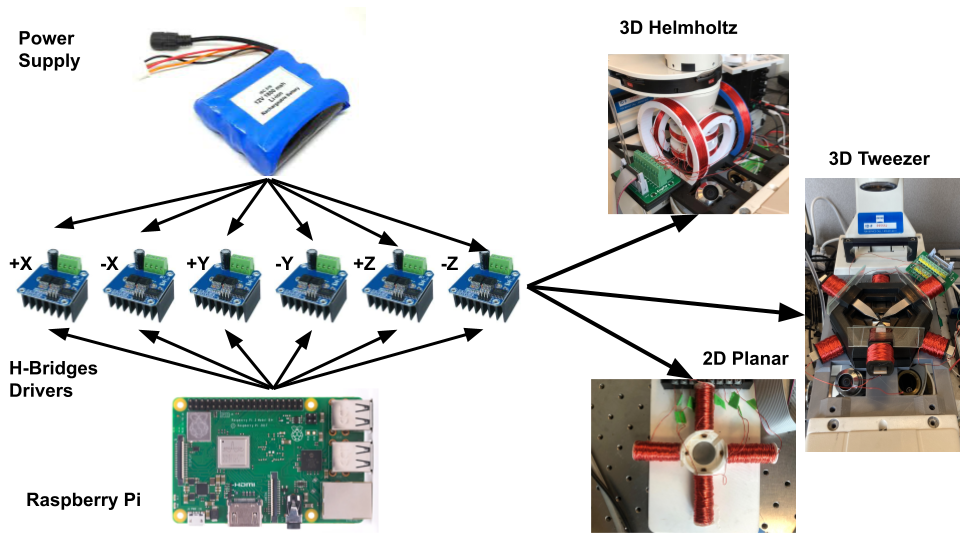}
    \caption{Electrical Schematic}
    \label{fig:ELE}
\end{figure}

The electrical system was designed to be functional with any and all of the above mentioned electromagnetic setups while also operating as a standalone device that is not constrained to a single lab computer. It is also fully portable and does not require a bench-top external power supply. The computational capability is provided by a Raspberry Pi Model 3 B+. The Raspberry Pi is a small computer and capable of running a complete operating system. The image file of Raspberry PI OS was burned to a SD card allowing a full Desktop experience when hooked up to an HDMI monitor or display. We employ a HMTECH 10.1 Inch Touchscreen Monitor with a resolution of 1024x600 to control the electromagnets. The touchscreen displays a custom graphical user interface which was written in Python. The computer is also equipped with 40 GPIO (general purpose input/output) pins which can be controlled through software. However, electromagnets require significantly high currents and are not able to be powered directly from the GPIO pins. Thus, six SEEU. AGAIN BTS7960B 43A Double DC Stepper H-Bridge PWM drivers are used to vary the current and polarity from an external power supply.  The H-Bridge convert the user defined currents generated from the GPIO pins and scale them accordingly to each electromagnet. The external power supply consists of 3 Lithium-ion 18650 batteries yielding 12V with a peak current draw of 3 A. A charging circuit board with a DC barrel jack allows the batteries to be recharged when empty. Power can also be supplied by an external power source if desired.  To connect the controller to each of the 3 coil setups, a 20 pin IDC cable with associated female adapter terminal was used.  The electrical schematic is shown in figure \ref{fig:ELE}.

\subsection{Software}
\subsubsection{Description}
In order to accurately control current to each coil in real time, a custom graphical user interface was designed in Python using the Tkinter library and Gpiozero libraries. The software enables application of constant magnetic fields in a user-defined direction as well as rolling, spinning, or vibrating functionalities. It also is capable of being controlled using a game controller for more responsiveness and precise control. 

To apply a constant magnetic field, for example for microrobot steering, the user-defined magnetic field direction and strength can be inputted and this is used to modulate the current going to each coil. The associated vector is broken into its x and y components and, since the magnetic field is proportional to the applied current, the current sent through the x and y coils is scaled proportionally. To enhance the field strength and uniformity, an opposite polarity signal is sent to the oppositely facing coil. For example, if a positive current is applied to the -x electromagnet coil, a simultaneous negative current will be applied to the +x coil. 

\begin{figure}[H]
    \centering
    \includegraphics[width=12cm]{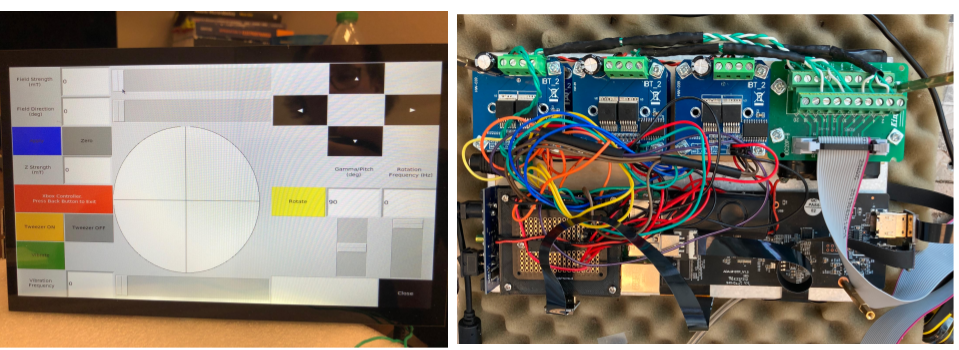}
    \caption{GUI python control panel displayed on 10 inch touchscreen. All circuitry mounted on backside of device.}
    \label{fig:GUI}
\end{figure}

For the use of magnetic tweezers, the user is able to toggle between "tweezer on" and "tweezer off" modes depending on whether the magnetic tweezers are being used or one of the other electromagnetic systems.  Magnetic tweezers concentrate the magnetic flux generated from traditional coils towards the tips of poles thereby generating adequate magnetic gradients capable of moving a magnetic micro-robot without any additional propulsion mechanism. To optimally move a microrobot, its magnetic moment should point in the same direction as the magnetic gradient. Therefore, when the magnetic tweezer mode is enabled in the software, current is only sent through one of the pairs of coils in order to align the magnetic moment with the field gradient. Note that the polarity of the magnetic field does not matter in this case since a single pole always acts to attract the microrobots. This is because changing the polarity changes the sign of $\textbf{B}$ and this also causes the microrobot to rotate such that $\textbf{m}$ aligns with B, hence changing the sign of $\textbf{m}$ as well.

All functions can also be controlled via a gaming controller. When a button is pressed on the graphical user interface, it will connect to a wireless USB gaming controller. The left joystick controls the 360$^{\circ}$  orientations or tweezer functionalities while the right joystick controls the rotating magnetic field direction, that is, the direction of a rolling magnetic microrobot. The device is also capable generating a rotating magnetic field in any user defined direction which allows for responsive and precise control of a rolling micro-robot or an artificial bacteria flagella micro-robot/helical micro-robot. This is done by applying a time varying sine wave to each of the X,Y,Z axis coils using the following equations:

\begin{equation}\label{Bx}
    B_x = A\cos(\gamma) \cos(\alpha) \cos(\omega t) + \sin(\alpha) \sin(\omega t)
\end{equation}
\begin{equation}\label{By}
    B_y = -A\cos(\gamma) \sin(\alpha) \cos(\omega t) + (\cos(\alpha) \sin(\omega t)
\end{equation}
\begin{equation}\label{Bz}
    B_z = A\sin(\gamma) \cos(\omega t
\end{equation}

\begin{figure}[H]
    \centering
    \includegraphics[width=12cm]{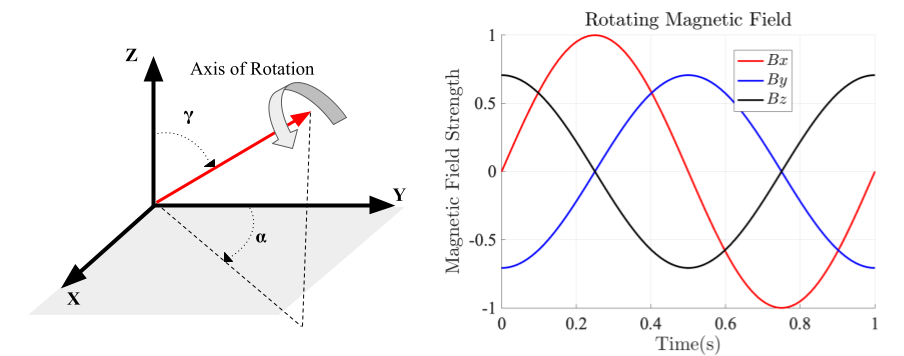}
    \caption{Left: Schematic illustrating the effect of $\alpha$ and $\gamma$ on rotation axis of microrobot. Right: A 1 Hz rotating magnetic field plotted using equations (\ref{Bx}),(\ref{By}), and (\ref{Bz}). 1 Hz sine waves with $\gamma$ = 45$^{\circ}$  and $\alpha$ = 90$^{\circ}$  over a 1 second interval.}
    \label{fig:Rot}
\end{figure}

Where $\gamma$ is the azimuthal angle from the Z axis, $\alpha$ is the polar angle from the Y axis, A is magnetic field magnitude and  $\omega$ is the frequency which controls the speed of the rolling microrobot.  When controlling magnetic rolling micro-robots, an aziumthal angle of 90$^{\circ}$  is set by default and the polar angle can be changed to steer the direction of the rolling micro-robot. See Figure \ref{fig:Rot} for an illustration of the rotation. For helical micro-robot swimmers, changing both the azimuthal and polar angle results in full 3D maneuverability. 

In addition to the functions described above, there are also buttons on the graphical user interface to magnetically vibrate a micro-robot. This is done by turning on and off opposite facing electromagnets at user defined frequencies. There are also quick direction buttons which allow for the application of magnetic fields in the $\pm$x or $\pm$y directions without the need to use the slider to select specific angles.

All of these functionalities can be controlled via the touchscreen display, or wirelessly via the PiGPIO daemon. This allows not only the GUI python program to be run and controlled on any other computer on the same Wi-Fi, but also opens up the possibility for integrating the magnetic control system with other software on the computer. For example, we have used this functionality for tracking microrobots using Python's OpenCV library and TrackPy while controlling the magnetic software wirelessly for real-time feedback control.. 

\section{Experimental Validation}

The 2D traditional coil setup is used to orient a variety of self-propelled microrobots including electrophoretic or diffusiophoretic magnetic Janus microspheres or bubble propelled microrobots. The 2D system was used in \cite{rivas2022dynamics} to study the dynamics and control of Platinum and Nickel coated bubble propelled micro-robots.

The 3D Helmholtz system is used to control a variety of magnetic rolling micro-robots i.e. micro-spheres coated with a hemisphere of Nickel via electron beam deposition. For example, the 3D Helmholtz system was used in \cite{rivas_cellular_nodate} to demonstrate cellular manipulation using rolling magnetic micro-robots. Furthermore, the system has also been used for the actuation of magnetic helical micro-robotic swimmers.

The magnetic tweezers can be used to direct magnetic micro-beads using high magnetic field gradients and is useful in a variety of applications where self-propulsion is not possible, or high forces are required. 
The 3D magnetic tweezers system was used in
\cite{mallick_doxorubicin-loaded_nodate} where 4.5 um paramagnetic micro-beads were coated with Doxorubicin and delivered via magnetic gradients to kill cancer cells.

\section{Conclusion}
This document described the design and construction of a compact, versatile, low cost, portable, and easy to use electromagnetic control system which can be universally paired with any of three electromagnetic systems: traditional planar coils, 3D magnetic tweezers, and 3D Helmholtz-based coils. Employing a graphical user interface allowed for straightforward control. We intend to continue making the system as robust and versatile as possible by incorporating other micro-robotic off board actuation methods such as acoustic and light actuation. We also envision that the portability of the system will make it useful for controlling microrobots in biomedical applications such as in medical imaging systems for in vitro or in vivo experiments. 

\backmatter

\section{Acknowledgements}
The authors gratefully acknowledge the late Richard West for his help with the cell lines. This project was supported by the Delaware INBRE program, with a grant from the National Institute of General Medical Sciences – NIGMS (P20 GM103446) from the National Institutes of Health and the State of Delaware. This work was also supported by NSF grant OIA2020973. This content is solely the responsibility of the authors and does not necessarily represent the official views of NIH.

\section*{Declaration of interests}

The authors declare that they have no known competing financial interests personal relationships that could have appeared to influence the work reported in this paper.

\bibliography{bibliography_template}
\end{document}